# A fast and simple algorithm for training neural probabilistic language models

**Andriy Mnih**                                                                                         AMNIH@GATSBY.UCL.AC.UK
Gatsby Computational Neuroscience Unit, University College London

**Yee Whye Teh**                                                                                        YWTEH@GATSBY.UCL.AC.UK
Gatsby Computational Neuroscience Unit, University College London

## Abstract

In spite of their superior performance, neural probabilistic language models (NPLMs) remain far less widely used than $n$-gram models due to their notoriously long training times, which are measured in weeks even for moderately-sized datasets. Training NPLMs is computationally expensive because they are explicitly normalized, which leads to having to consider all words in the vocabulary when computing the log-likelihood gradients.

We propose a fast and simple algorithm for training NPLMs based on noise-contrastive estimation, a newly introduced procedure for estimating unnormalized continuous distributions. We investigate the behaviour of the algorithm on the Penn Treebank corpus and show that it reduces the training times by more than an order of magnitude without affecting the quality of the resulting models. The algorithm is also more efficient and much more stable than importance sampling because it requires far fewer noise samples to perform well.

We demonstrate the scalability of the proposed approach by training several neural language models on a 47M-word corpus with a 80K-word vocabulary, obtaining state-of-the-art results on the Microsoft Research Sentence Completion Challenge dataset.

## 1. Introduction

By assigning probabilities to sentences, language models allow distinguishing between probable and improbable sentences, which makes such models an important component of speech recognition, machine translation, and information retrieval systems. Probabilistic language models are typically based on the Markov assumption, which means that they model the conditional distribution of the next word in a sentence given some fixed number of words that immediately precede it. The group of words conditioned on is called the *context*, denoted $h$, while the word being predicted is called the *target* word, denoted $w$. $n$-gram models, which are effectively smoothed tables of normalized word/context co-occurrence counts, have dominated language modelling for decades due to their simplicity and excellent performance.

In the last few years neural language models have become competitive with $n$-grams and now routinely outperform them (Mikolov et al., 2011). NPLMs model the distribution for the next word as a smooth function of learned multi-dimensional real-valued representations of the context words and the target word. Similar representations are learned for words that are used in similar ways, ensuring that the network outputs similar probability values for them. Word representations learned by language models are also used for natural language processing applications such as semantic role labelling (Collobert & Weston, 2008), sentiment analysis (Maas & Ng, 2010), named entity recognition (Turian et al., 2010), and parsing (Socher et al., 2011).

Unfortunately, NPLMs are very slow to train, which makes them unappealing for large-scale applications. This is a consequence of having to consider the entire vocabulary when computing the probability of a single word or the corresponding gradient. In fact, the time complexity of this computation scales as the product of the vocabulary size and the word feature dimensionality. One way to accelerate this computation is to reduce the vocabulary size for the NPLM by using it to predict only the most frequent words and handling the rest using an $n$-gram model (Schwenk & Gauvain, 2005).

Alternatively, the vocabulary can be structured into a tree with words at the leaves, allowing exponentially faster computation of word probabilities and their gradients (Morin & Bengio, 2005). Unfortunately, the predictive performance of the resulting model is heavily dependent on the tree used and finding a good tree is a difficult problem (Mnih & Hinton, 2009).

Perhaps a more elegant approach is to keep the model the same and to approximate the expensive gradient computations using importance sampling (Bengio & Senécal, 2003). Unfortunately, the variance in the importance sampling estimates can make learning unstable, unless it is carefully controlled (Bengio & Senécal, 2008).

In this paper we propose an efficient algorithm for training NPLMs based on noise-contrastive estimation (Gutmann & Hyvärinen, 2010), which is much more stable than importance sampling. Though it also uses sampling to approximate the gradients needed for learning, neither the number of samples nor the proposal distribution require dynamic adaptation for achieving performance on par with maximum likelihood learning.

## 2. Neural probabilistic language models

A statistical language model is simply a collection of conditional distributions for the next word, indexed by its context.[1] In a neural language model the conditional distribution corresponding to context $h$, $P^h(w)$, is defined as

$$P_\theta^h(w) = \frac{\exp(s_\theta(w, h))}{\sum_{w'} \exp(s_\theta(w', h))}, \quad (1)$$

where $s_\theta(w, h)$ is the scoring function with parameters $\theta$ which quantifies the compatibility of word $w$ with context $h$. The negated scoring function is sometimes referred to as the energy function (Bengio et al., 2000).

Depending on the form of $s_\theta(w, h)$, Eq. 1 can describe both neural and maximum entropy language models (Berger et al., 1996). The main difference between these two model classes lies in the features they use: neural language models learn their features jointly with other parameters, while maximum entropy models use fixed hand-engineered features and only learn the weights for those features. A neural language model represents each word in the vocabulary using a real-valued feature vector and defines the scoring function in terms of the feature vectors of the context words and the next word. In some models, different feature vector tables are used for the context and the next word vocabularies (Bengio et al., 2000), while in others the table is shared (Bengio et al., 2003; Mnih & Hinton, 2007).

---

[1] Though almost all statistical language models predict the next word, it is also possible to model the distribution of the word preceding the context or surrounded by the context.

The feature vectors account for the vast majority of parameters in neural language models, which means that their memory requirements are linear in the vocabulary size. This compares favourably to the memory requirements of the $n$-gram models, which are typically linear in the training set size.

### 2.1. Log-bilinear model

The training method we propose is directly applicable to all neural probabilistic and maximum-entropy language models. For simplicity, we will perform our experiments using the log-bilinear language (LBL) model (Mnih & Hinton, 2007), which is the simplest neural language model. The LBL model performs linear prediction in the semantic word feature space and does not have non-linearities. In spite of its simplicity, the LBL model has been shown to outperform $n$-grams, though the more complex neural language models (Mikolov et al., 2010; Mnih et al., 2009) can outperform it.

In this paper we will use a slightly extended version of the LBL model that uses separate feature vector tables for the context words and the target words. Thus a context word $w$ will be represented with feature vector $r_w$, while a target word $w$ will be represented with feature vector $q_w$. Given a context $h$, the model computes the predicted representation for the target word by linearly combining the feature vectors for the context words using position-dependent context weight matrices $C_i$:

$$\hat{q} = \sum_{i=1}^{n-1} C_i r_{w_i}. \quad (2)$$

The score for the match between the context and the next word is computed by taking the dot product between the predicted representation and the representation of the candidate target word $w$:

$$s_\theta(w, h) = \hat{q}^\top q_w + b_w. \quad (3)$$

Here $b_w$ is the base rate parameter used to model the popularity of $w$. The probability of $w$ in context $h$ is then obtained by plugging the above score function into Eq. 1.

## 2.2. Maximum likelihood learning

Maximum likelihood training of neural language models is tractable but expensive because computing the gradient of log-likelihood takes time linear in the vocabulary size. The contribution of a single context/word observation to the gradient of the log-likelihood is given by

$$\frac{\partial}{\partial \theta} \log P_\theta^h(w) = \frac{\partial}{\partial \theta} s_\theta(w, h) - \sum_{w'} P_\theta^h(w') \frac{\partial}{\partial \theta} s_\theta(w', h) \quad (4)$$

$$= \frac{\partial}{\partial \theta} s_\theta(w, h) - E_{P_\theta^h}\left[\frac{\partial}{\partial \theta} s_\theta(w, h)\right].$$

The expectation w.r.t. $P_\theta^h(w')$ is expensive to evaluate because it requires computing $s_\theta(w, h)$ for all words in the vocabulary. Since vocabularies typically contain tens of thousands of words, maximum likelihood learning tends to be very slow.

## 2.3. Importance sampling

Bengio and Senécal (2003) have proposed a method for speeding up training of neural language models based on approximating the expectation in Eq. 4 using importance sampling. The idea is to generate $k$ samples $x_1, ..., x_k$ from an easy-to-sample-from distribution $Q^h(w)$ and estimate the gradient with

$$\frac{\partial}{\partial \theta} \log P_\theta^h(w) \approx \frac{\partial}{\partial \theta} s_\theta(w, h) - \frac{1}{V} \sum_{j=1}^{k} v(x_j) \frac{\partial}{\partial \theta} s_\theta(x_j, h), \quad (5)$$

where $v(x) = \frac{\exp(s_\theta(x,h))}{Q^h(w=x)}$ and $V = \sum_{j=1}^{k} v(x_j)$. The normalization by $V$ is necessary here because the importance weights $v$ are computed using the unnormalized model distribution $\exp(s_\theta(x, h))$. Typically the proposal distribution is an $n$-gram model fit to the training set, possibly with a context size different from the neural model's.

Though this approach is conceptually simple, it is non-trivial to use in practice because the high variance of the importance sampling estimates can make learning unstable. The variance tends to grow as learning progresses, because the model distribution moves away from the proposal distribution.[2] One way to control the variance is to keep increasing the number of samples during training so that the effective sample size stays above some predetermined value (Bengio & Senécal, 2003). Alternatively, the $n$-gram proposal distribution can be adapted to track the model distribution throughout training (Bengio & Senécal, 2008). The first approach is simpler but less efficient because the increasing number of samples makes learning slower. The second approach leads to greater speedups but is considerably more difficult to implement and requires additional memory for storing the adaptive proposal distribution.

## 3. Noise-contrastive estimation

We propose using noise-contrastive estimation (NCE) as a more stable alternative to importance sampling for efficient training of neural language models and other models defined by Eq. 1. NCE has recently been introduced by Gutmann and Hyvärinen (2010) for training unnormalized probabilistic models. Though it has been developed for estimating probability densities, we are interested in applying it to discrete distributions and so will assume discrete distributions and use probability mass functions instead of density functions.

The basic idea of NCE is to reduce the problem of density estimation to that of binary classification, discriminating between samples from the data distribution and samples from a known noise distribution. In the language modelling setting, the data distribution $P_d^h(w)$ will be the distribution of words that occur after a particular context $h$. Though it is possible to use context-dependent noise distributions, for simplicity we will use a context-independent (unigram) $P_n(w)$. We are interested in fitting the context-dependent model $P_\theta^h(w)$ to $P_d^h(w)$.

Following Gutmann and Hyvärinen (2012), we assume that noise samples are $k$ times more frequent than data samples, so that datapoints come from the mixture $\frac{1}{k+1} P_d^h(w) + \frac{k}{k+1} P_n(w)$. Then the posterior probability that sample $w$ came from the data distribution is

$$P^h(D=1|w) = \frac{P_d^h(w)}{P_d^h(w) + kP_n(w)}. \quad (6)$$

Since we would like to fit $P_\theta^h$ to $P_d^h$, we use $P_\theta^h$ in place of $P_d^h$ in Eq. 6, making the posterior probability a function of the model parameters $\theta$:

$$P^h(D=1|w,\theta) = \frac{P_\theta^h(w)}{P_\theta^h(w) + kP_n(w)}. \quad (7)$$

---

[2] Bengio and Senécal (2008) argue that this happens because neural language models and $n$-gram models learn very different distributions.

This quantity can be too expensive to compute, however, because of the normalization required for evaluating $P_\theta^h(w)$ (Eq. 1). NCE sidesteps this issue by avoiding explicit normalization and treating normalization constants as parameters. Thus the model is parameterized in terms of an unnormalized distribution $P_{\theta^0}^{h0}$ and a learned parameter $c^h$ corresponding to the logarithm of the normalizing constant:

$$P_\theta^h(w) = P_{\theta^0}^{h0}(w)\exp(c^h). \qquad (8)$$

Here $\theta^0$ are the parameters of the unnormalized distribution and $\theta = \{\theta^0, c^h\}$.

To fit the context-dependent model to the data (for the moment ignoring the fact that it shares parameters with models for other contexts), we simply maximize the expectation of $\log P^h(D|w, \theta)$ under the mixture of the data and noise samples. This leads to the objective function

$$J^h(\theta) = E_{P_d^h}\left[\log \frac{P_\theta^h(w)}{P_\theta^h(w) + kP_n(w)}\right] + \qquad (9)$$
$$kE_{P_n}\left[\log \frac{kP_n(w)}{P_\theta^h(w) + kP_n(w)}\right]$$

with the gradient

$$\frac{\partial}{\partial \theta}J^h(\theta) = E_{P_d^h}\left[\frac{kP_n(w)}{P_\theta^h(w) + kP_n(w)}\frac{\partial}{\partial \theta}\log P_\theta^h(w)\right] - \qquad (10)$$
$$kE_{P_n}\left[\frac{P_\theta^h(w)}{P_\theta^h(w) + kP_n(w)}\frac{\partial}{\partial \theta}\log P_\theta^h(w)\right].$$

Note that the gradient can also be expressed as

$$\frac{\partial}{\partial \theta}J^h(\theta) = \sum_w \frac{kP_n(w)}{P_\theta^h(w) + kP_n(w)} \times \qquad (11)$$
$$(P_d^h(w) - P_\theta^h(w))\frac{\partial}{\partial \theta}\log P_\theta^h(w),$$

and that as $k \to \infty$,

$$\frac{\partial}{\partial \theta}J^h(\theta) \to \sum_w (P_d^h(w) - P_\theta^h(w))\frac{\partial}{\partial \theta}\log P_\theta^h(w), \qquad (12)$$

which is the maximum likelihood gradient. Thus as the ratio of noise samples to observations increases, the NCE gradient approaches the maximum likelihood gradient.

In practice, given a word $w$ observed in context $h$, we compute its contribution to the gradient by generating $k$ noise samples $x_1, ..., x_k$ and using the formula

$$\frac{\partial}{\partial \theta}J^{h,w}(\theta) = \frac{kP_n(w)}{P_\theta^h(w) + kP_n(w)}\frac{\partial}{\partial \theta}\log P_\theta^h(w) - \qquad (13)$$
$$\sum_{i=1}^{k}\left[\frac{P_\theta^h(x_i)}{P_\theta^h(x_i) + kP_n(x_i)}\frac{\partial}{\partial \theta}\log P_\theta^h(x_i)\right].$$

Note that the weights $\frac{P_\theta^h(x_i)}{P_\theta^h(x_i) + kP_n(x_i)}$ are always between 0 and 1, which makes NCE-based learning very stable (Gutmann & Hyvärinen, 2010). In contrast, the weights produced by importance sampling can be arbitrarily large.

Since the distributions for different contexts share parameters, we cannot learn these distributions independently of each other by optimizing one $J^h(\theta)$ at a time. Instead, we define a global NCE objective by combining the per-context NCE objectives using the empirical context probabilities $P(h)$ as weights:

$$J(\theta) = \sum_h P(h)J^h(\theta). \qquad (14)$$

Note that this is the traditional approach for combining the per-context ML objectives for training neural language models.

### 3.1. Dealing with normalizing constants

Our initial implementation of NCE training learned a (log-)normalizing constant ($c$ in Eq. 8) for each context in the training set, storing them in a hash table indexed by the context.[3] Though this approach works well for small datasets, it requires estimating one parameter per context, making it difficult to scale to huge numbers of observed contexts encountered by models with large context sizes. Surprisingly, we discovered that fixing the normalizing constants to 1,[4] instead of learning them, did not affect the performance of the resulting models. We believe this is because the model has so many free parameters that meeting the approximate per-context normalization constraint encouraged by the objective function is easy.

### 3.2. Potential speedup

We will now compare the gradient computation costs for NCE and ML learning. Suppose $c$ is the context size, $d$ is the word feature vector dimensionality, and $V$ is the vocabulary size of the model. Then computing the predicted representation using Eq. 2 takes about $cd^2$ operations for both NCE and ML. For ML, computing the distribution of the next word from the predicted representation takes about $Vd$ operations. For NCE, evaluating the probability of $k$ noise samples under the model takes about $kd$ operations. Since the gradient computation in each model has the same complexity as computing the probabilities, the speedup for

---

[3] We did not use the learned normalizing constants when computing the validation and test set perplexities. Rather we normalized the probabilities explicitly.

[4] This amounts to setting the normalizing parameters $c$ to 0.

Table 1. Results for the LBL model with 100D feature vectors and a 2-word context on the Penn Treebank corpus.

| Training algorithm | Number of samples | Test PPL | Training time (h) |
|---|---|---|---|
| ML |  | 163.5 | 21 |
| NCE | 1 | 192.5 | 1.5 |
| NCE | 5 | 172.6 | 1.5 |
| NCE | 25 | 163.1 | 1.5 |
| NCE | 100 | 159.1 | 1.5 |

Table 2. The effect of the noise distribution and the number of noise samples on the test set perplexity.

| Number of samples | PPL using unigram noise | PPL using uniform noise |
|---|---|---|
| 1 | 192.5 | 291.0 |
| 5 | 172.6 | 233.7 |
| 25 | 163.1 | 195.1 |
| 100 | 159.1 | 173.2 |

each parameter update due to using NCE is about

$$\text{Speedup} = \frac{cd^2 + Vd}{cd^2 + kd} = \frac{cd + V}{cd + k}. \quad (15)$$

For a model with a 2-word context, 100D feature vectors, and a vocabulary size of 10K, an NCE update using 25 noise samples should be about 45 times faster than an ML update.

Since the time complexity of computing the predicted representation is quadratic in the feature vector dimensionality, it can dominate the cost of the parameter update, making learning slow even for a small number of noise samples. We can avoid this by making context weight matrices $C_i$ diagonal, reducing the complexity of computing the predicted representation to $cd$, and making the speedup factor $\frac{c+V}{c+k}$. For the model above this would amount to a factor of 370. The use of diagonal context matrices was introduced by Mnih & Hinton (2009) to speed up their hierarchical LBL-like model.

Since the cost of a parameter update for importance-sampling-based learning is the same as for NCE with the same number of noise samples, the algorithm that needs fewer samples to perform well will be faster.

## 4. Penn Treebank results

We investigated the properties of the proposed algorithm empirically on the Penn Treebank corpus. As is common practice, we trained on sections 0-20 (930K words) and used sections 21-22 (74k words) as the validation set and sections 23-24 (82k words) as the test set. The standard vocabulary of 10K most frequent words was used with the remaining words replaced by a special token. We chose to use this dataset to keep the training time for exact maximum likelihood learning reasonable.

The learning rates were adapted at the end of each epoch based on the change in the validation set perplexity since the end of the previous epoch. The rates were halved when the perplexity increased and were left unchanged otherwise. Parameters were updated based on mini-batches of 1000 context/word pairs each. Except when stated otherwise, NCE training generated 25 noise samples from the empirical unigram distribution per context/word observation. Noise samples were generated anew for each update. We did not use a weight penalty as the validation-score-based learning rate reduction appeared to be sufficient to avoid overfitting. All models used a two-word context and different 100D feature vector tables for context and target words.

Our first experiment compared ML learning to NCE learning for various numbers of noise samples. The resulting test perplexities and training times are shown in Table 1. It is clear that increasing the number of noise samples produces better-performing models, with 25 samples being sufficient to match the ML-trained model. In terms of training time, NCE was 14 times faster than ML. The number of noise samples did not have a significant effect on the running time because computing the predicted representation was considerably more expensive than computing the probability of (at most) 100 samples. The main reason the speedup factor was less than 45 (the value predicted in Section 3.2) is because NCE took about twice as many epochs as ML to converge. Our NCE implementation is also less optimized than the ML implementation which takes greater advantage of the BLAS matrix routines.

To explore the effect of the noise distribution on the performance of the algorithm, we tried generating noise samples from the unigram as well as the uniform distribution. For each noise distribution we trained models using 1, 5, 25, and 100 noise samples per datapoint. As shown in Table 2, the unigram noise distribution leads to much better test set perplexity in all cases. However, the perplexity gap shrinks as the number of noise samples increases, from almost 100 for a single noise sample down to under 15 for 100 noise samples. In spite of poor test set performance, a uniform noise distribution did not lead to learning instability even when a single noise sample was used.

In addition to the ML and NCE algorithms, we also implemented the importance sampling training algorithm from (Bengio & Senécal, 2003) to use as a baseline, but found it very unstable. It diverged in virtually all of our experiments, even with adaptive sample size and the target effective sample size set to hundreds. The only run that did not diverge involved learning a unigram model using the target unigram as the proposal distribution, which is the ideal situation for importance sampling. The cause of failure in all cases was the appearance of extremely large importance weights once the model distribution became sufficiently different from the unigram proposal distribution[5], which is a known problem with importance sampling. Since IS-based methods seem to require well over a hundred samples per gradient computation (Bengio & Senécal, 2008), even when an adaptive proposal distribution is used, we expect IS-based training to be considerably slower than NCE, which, as we have shown, can achieve ML-level performance with only 25 noise samples.

## 5. Sentence Completion Challenge

To demonstrate the scalability and effectiveness of our approach we used it to train several large neural language models for the Microsoft Research Sentence Completion Challenge (Zweig & Burges, 2011). The challenge was designed as a benchmark for semantic models and consists of SAT-style sentence completion problems. Given 1,040 sentences, each of which is missing a word, the task is to select the correct word out of the five candidates provided for each sentence. Candidate words have been chosen from relatively infrequent words using a maximum entropy model to ensure that the resulting complete sentences were not too improbable. Human judges then picked the best four candidates for each sentence so that all completions were grammatically correct but the correct answer was unambiguous. Though humans can achieve over 90% accuracy on the task, statistical models fare much worse with the best result of 49% produced by a whole-sentence LSA model, and $n$-gram models achieving only about 39% accuracy (Zweig & Burges, 2011).

Neural language models are a natural choice for this task because they can take advantage of larger contexts than traditional $n$-gram models, which we expect to be important for sentence completion. We used a slightly modified LBL architecture for our models for this task. In the interests of scalability, we used diagonal context weight matrices which reduced the time complexity of gradient computations from quadratic to linear in the dimensionality of word feature vectors and allowed us to use more feature dimensions. Since the task was sentence completion, we made the models aware of sentence boundaries by using a special "out-of-sentence" token for words in context positions outside of the sentence containing the word being predicted. For example, this token would be used as the context word when predicting the first word in a sentence using a model with a single-word context.

---

[5]Though using a unigram proposal distribution might appear naive, Bengio and Senécal (2003) reported that higher-order $n$-gram proposal distributions worked much worse than the unigram.

We score a candidate sentence with a language model by using it to compute the probability of each word in the sentence and taking the product of those probabilities as the sentence score. We then pick the candidate word that produces the highest-scoring sentence as our answer. Note that this way of using a model with a $c$-word context takes into account $c$ words on *both* sides of the candidate word because the probabilities of the $c$ words following the candidate word depend on it.

The models were trained on the standard training set for the challenge containing 522 works from Project Gutenberg. After removing the Project Gutenberg headers and footers from the files, we split them into sentences and then tokenized the sentences into words. We used the Punkt sentence tokenizer and the Penn Treebank word tokenizer from NLTK (Bird et al., 2009). We then converted all words to lowercase and replaced the ones that occurred fewer than 5 times with an "unknown word" token, resulting in a vocabulary size of just under 80,000. The sentences to be completed were preprocessed in the same manner. The resulting dataset was then randomly split at the sentence level into a test and validation sets of 10K words (500 sentences) each and a 47M-word training set.

We used the training procedure described in Section 4, with the exception of using a small weight penalty to avoid overfitting. Each model took between one and two days to train on a single core of a modern CPU. As a baseline for comparison, we also trained several $n$-gram models (with modified Kneser-Ney smoothing) using the SRILM toolkit (Stolcke, 2002), obtaining results similar to those reported by Zweig & Burges (2011).

Since we selected hyperparameters based on the (Gutenberg) validation set perplexity, we report the scores on the entire collection of 1,040 sentences, which means that our results are directly comparable to those of Zweig & Burges (2011). As can be seen from Ta-ble 3, more word features and larger context leads to better performance in LBL models in terms of both accuracy and perplexity. The LBL models perform considerably better on sentence completion than $n$-gram models, in spite of having higher test perplexity. Even the LBL model with a two-word context performs better than any $n$-gram model. The LBL model with a five-word context, matches the best published result on the dataset. Note that the LSA model that produced that result considered all words in a sentence, while an LBL model with a $c$-word contexts considers only the $2c$ words that surround the candidate word. The model with a 10-word context and 300D feature vectors outperforms the LSA model by a large margin and sets a new accuracy record for the dataset at 54.7%.

*Table 3.* Accuracy on the complete MSR Sentence Completion Challenge dataset. $n \times 2$ indicates a bidirectional context. The LSA result is from (Zweig & Burges, 2011).

| MODEL TYPE | CONTEXT SIZE | LATENT DIM | PERCENT CORRECT | TEST PPL |
|---|---|---|---|---|
| 3-GRAM | 2 | | 36.0 | 130.8 |
| 4-GRAM | 3 | | 39.1 | 122.1 |
| 5-GRAM | 4 | | 38.7 | 121.5 |
| 6-GRAM | 5 | | 38.4 | 121.7 |
| LSA | SENTENCE | 300 | 49 | |
| LBL | 2 | 100 | 41.5 | 145.5 |
| LBL | 3 | 100 | 45.1 | 135.6 |
| LBL | 5 | 100 | 49.3 | 129.8 |
| LBL | 10 | 100 | 50.0 | 124.0 |
| LBL | 5 | 200 | 50.8 | 123.6 |
| LBL | 10 | 200 | 52.8 | 117.7 |
| LBL | 10 | 300 | 54.7 | 116.4 |
| LBL | 10×2 | 100 | 44.5 | 38.6 |
| LBL | 10×2 | 200 | 49.8 | 33.6 |

Language models typically use the words preceding the word of interest as the context. However, since we are interested in filling in a word in the middle of the sentence, it makes sense to use both the preceding and the following words as the context for the language model, making the context bidirectional. We trained several LBL models with bidirectional context to see whether such models are superior to their unidirectional-context counterparts for sentence completion. Scoring a sentence with a bidirectional model is both simpler and faster: we simply compute the probability of the candidate word under the model using the context surrounding the word. Thus a model is applied only once per sentence, instead of $c+1$ times required by the unidirectional models.

As Table 3 shows, the LBL models with bidirectional contexts achieve much lower test perplexity than their unidirectional counterparts, which is not surprising because they also condition on words that follow the word being predicted. What is surprising, however, is that bidirectional contexts appear to be considerably less effective for sentence completion than unidirectional contexts. Though the $c$-word context model and $c \times 2$-word context model look at the same words when using the scoring procedures we described above, the unidirectional model seems to make better use of the available information.

## 6. Discussion

We have introduced a simple and efficient method for training statistical language models based on noise-contrastive estimation. Our results show that the learning algorithm is very stable and can produce models that perform as well as the ones trained using maximum likelihood in less than one-tenth of the time. In a large-scale test of the approach, we trained several neural language models on a collection of Project Gutenberg texts, achieving state-of-the-art performance on the Microsoft Research Sentence Completion Challenge dataset.

Though we have shown that the unigram noise distribution is sufficient for training neural language models efficiently, context-dependent noise distributions are worth investigating because they might lead to even faster training by reducing the number of noise samples needed.

Recently, Pihlaja et al. (2010) introduced a family of estimation methods for unnormalized models that includes NCE and importance sampling as special cases. Other members of this family might be of interest for training language models, though our preliminary results suggest that none of them outperform NCE.

Finally, we believe that NCE can be applied to many models other than neural or maximum-entropy language models. Probabilistic classifiers with many classes are a prime candidate.

## Acknowledgments

We thank Vinayak Rao and Lloyd Elliot for their helpful comments. We thank the Gatsby Charitable Foundation for generous funding.